\documentclass[11pt,a4paper]{article}
\usepackage[hyperref]{naaclhlt2018}
\usepackage{times}
\usepackage{latexsym}
\usepackage{graphicx}
\usepackage[utf8]{inputenc} 
\usepackage[T1]{fontenc} 
\usepackage{multirow}

\usepackage{url}

\aclfinalcopy 

\title{NL-FIIT at SemEval-2019 Task 9:  \\ Neural Model Ensemble for Suggestion Mining}

\author{
  Samuel Pecar, Marian Simko, Maria Bielikova \\
  Slovak University of Technology in Bratislava \\ 
  Faculty of Informatics and Information Technologies \\ 
  Ilkovicova 2, 842 16 Bratislava, Slovakia \\
  {\tt \{samuel.pecar, marian.simko, maria.bielikova\}@stuba.sk} \\
}

\date{}

\begin{document}
\maketitle
\begin{abstract}
In this paper, we present neural model architecture submitted to the SemEval-2019 Task 9 competition: "Suggestion Mining from Online Reviews and Forums". We participated in both subtasks for domain specific and also cross-domain suggestion mining. We proposed a recurrent neural network architecture that employs Bi-LSTM layers and also self-attention mechanism. Our architecture tries to encode words via word representations using ELMo and ensembles multiple models to achieve better results. 
We performed experiments with different setups of our proposed model involving weighting of prediction classes for loss function. Our best model achieved in official test evaluation score of 0.6816 for subtask A and 0.6850 for subtask B. In official results, we achieved 12th and 10th place in subtasks A and B, respectively.
\end{abstract}

\section{Introduction}

Review-based portals and online forums contain plethora of user-generated text. We can consider customer reviews and inputs from online forums as an important source of novel information. These texts often contain many different opinions, which are the subject of research in area of opinion mining. 

On the other hand, there can be also different types of information within these texts, such as suggestions. Unlike opinions, suggestions can appear in different parts of text and also appear more sparsely. Suggestion mining, as defined in this task, can be realized as standard text classification. We perform classification to two classes, which are suggestion and non-suggestion. 

As presented by organizers, suggestion mining has different challenges \cite{Negi2019semeval}:
\begin{itemize}
    \item Class imbalance - suggestions appear very sparsely in reviews and forums and most of the samples are negatively sampled,
    \item Figurative expressions - expression can be often found in social networks but it is not always in form of suggestion,
    \item Context dependency - some sentences can be viewed as a suggestion, if it appears in specific domain or surrounded by specific sentences,
    \item Long and complex sentences - suggestions can be expressed as only small part of original sentence, which can be much longer.
\end{itemize}

Unlike opinions, suggestions can be more likely extracted also by pattern matching. We can extract suggestions by different heuristic features and keywords, such as \textit{suggest, recommend, advise} \cite{Negi2015}. Some works deal with domain terminology, thesaurus, linguistic parser and extraction rules \cite{Brun2013suggestion}. Linguistic rules were also used for identification and extraction in sentiment expression \cite{Viswanathan2011suggestion}.

We believe that different extracted information from customer reviews and online forums can offer a valuable input for both customers and owners of products or forums and this information can be also a subject for automatic opinion summarization \cite{Pecar2018summarization}.

In this paper, we present a neural network architecture consisting of different types of layers, such as embedding, recurrent, transformer or self-attention layer. We continued in our previous work on multi-level pre-processing \cite{Pecar2018wassa}. We performed experiments with different word representations, such as ELMo, BERT or GloVe. We report results of our experiments along with error analysis of our models.

\section{Model}

\begin{figure}[ht]
\centerline{\includegraphics[scale=0.32]{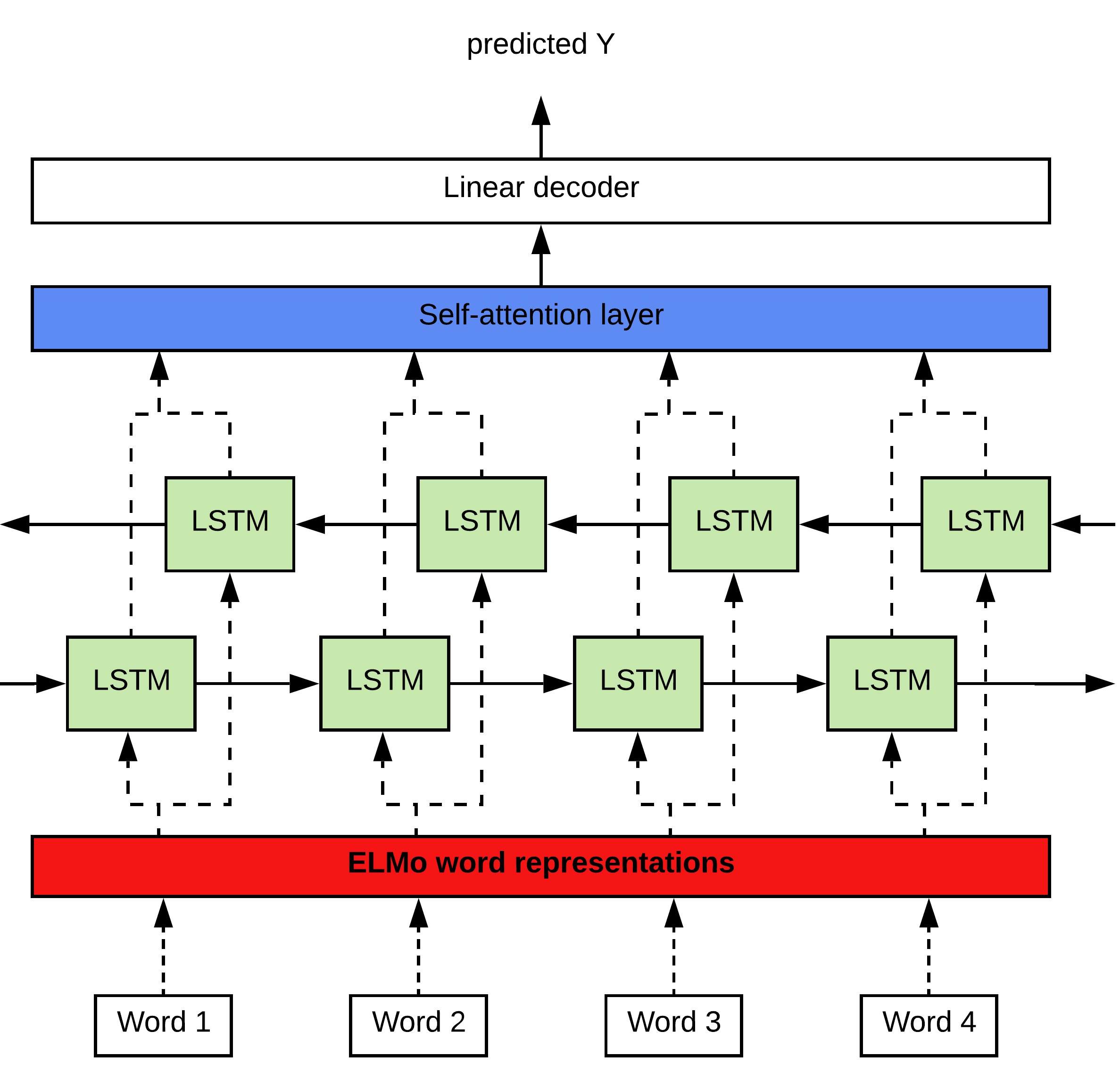}}
\caption{Proposed neural model architecture}
\label{fig:model}
\end{figure}

In this SemEval task, we experimented with multiple setups based on different types of neural layers on the top of an embedding layer. In Figure \ref{fig:model}, we show general architecture of our proposed model. We also experimented with a transformer encoder, which is described in the paragraph on encoder layer below. 

\paragraph{Preprocessing}

We consider preprocessing of input samples as one of the most important phases in natural language processing. For user generated content, preprocessing is even more important due to noisy and ungrammatical text. We performed a study on the impact of preprocessing in our previous work \cite{Pecar2018wassa}. 

For this suggestion mining task, we used preprocessing in several stages, which were performed in order as follows:
\begin{enumerate}
    \item text cleaning -- removing all characters from not Latin alphabet, such as Cyrillic, Greek or Chinese characters,
    \item character and word normalization -- normalization of different use of characters and words, such as apostrophe, punctuation, date and time (e.g. \textit{‘’} for apostrophe and \textit{”“„} for quotation),
    \item shorten phrase expanding -- expanding all shorten phrases to their appropriate long form (e.g. \textit{I'll} to \textit{I will}),
    \item expanding negations -- expanding all negation forms, which appeared in short form to their appropriate long form (e.g. \textit{aren't, won't} to \textit{are not, will not}),
    \item punctuation escaping -- escaping all punctuation with spaces do separate those characters from words.
\end{enumerate}

\paragraph{Word Representations}
To represent words from samples, we used deep contextualized word representations \cite{Peters2018} also known as ELMo along with its available pre-trained model\footnote{https://allennlp.org/elmo}. We also experimented with transformers model word representations known as BERT \cite{Devlin2018bert} and its pre-trained model\footnote{https://github.com/google-research/bert}. For language modeling in subtask B, we also experimented with GloVe embeddings \cite{Pennington2014}.

\paragraph{Encoder layer}

Word representations are fed into different encoder layers. Mostly, we used different setups of Bi-LSTM. We experimented with multiple stacked recurrent layers with different number of units within layers. In both cases (only one layer, multiple stacked layers) we used also self-attention mechanism to improve results and reduce over-fitting to the train dataset. We also tried to experiment with different attention layers and used transformer encoder \cite{Vaswani2017} but due to very high requirements for memory, we were not able to run model with full size of this network and smaller networks produced significantly worse results.

\paragraph{Decoder Layer}

We used standard linear layer to decode output representation of recurrent layers with self-attention mechanism to class probabilities. In case of model ensemble, we needed to employ also another logarithmic softmax function for better interpretations of probabilities of samples for both classes.

\paragraph{Loss Function}
For a loss function, we experimented with the standard cross-entropy loss and the negative log likelihood loss in case a logarithmic softmax were used. We also experimented with weight setup for classes for loss contribution.

\paragraph{Model Ensemble}

Model ensemble can be considered as a useful technique to obtain better results than using only single model for predictions. We experimented with different size of model ensemble and also two different types. In one model, we tried averaging model prediction probabilities and in second one, we used voting mechanism and predicted class with more votes.

\paragraph{Regularization}
To reduce possibilities of over-fitting to train dataset, we used also dropout as a regularization technique. We used dropout on embedding layer output, on encoder output along with dropout between stacked RNN layers and also at the output of attention layer. We used different dropout probabilities in range from 0.2 to 0.6. 

\section{Evaluation}
In this section, we briefly summarize basic information about used dataset. Later, we describe different setups of our model. Each team could submit in total 4 submission as an official results. For evaluation, binary F1 measure (F1 score over positive labels) was taken as an official results of submission.  

\subsection{Dataset}

The dataset for suggestion mining task consists of feedback posts on Universal Windows Platform available on \textit{uservoice.com}. The dataset contains only labels for two categories: the text is suggestion or it is not. The train dataset contains approximately 9 thousands of text samples. For validation, there were available approximately 600 samples for subtask A and 800 samples for subtask B. The size of test datasets were approximately 800 and 1000 samples for subtask A and B, respectively. More detailed information can be found in the main paper of the task \cite{Negi2019semeval}.

\subsection{Results}

In Table \ref{tab:official}, we provide basic information about setups of performed submission. For every setup, we used ELMo word representations as an embedding layer, different setups of dropout in each layer of neural network in the range from 0.3 to 0.6. Each LSTM layer has its hidden size set to 1024 units per layer. For some submissions, we also experimented with model ensemble. We took different number of models, which had the best performance on development (trial) set and used averaging predicted probabilities to get final prediction or voting mechanism and get label with more votes. In subtask A, we used 5 best trained models for voting model ensemble and 3 models for mean model ensemble. In subtask B, we used 3 best trained models for model ensemble. 

\begin{table*}[ht]
\begin{center}
\begin{tabular}{c|c|c|c|r|r|r}
\multirow{2}{*}{\bf task}  & 
\multirow{2}{*}{\bf submission} & 
\multirow{2}{*}{\bf layers} &
\multicolumn{1}{c|}{\bf model} &
\multirow{2}{*}{\bf micro F1} & 
\multirow{2}{*}{\bf macro F1} & 
\multirow{2}{*}{\bf binary F1}  \\
 &  &  & \bf ensemble &  &  & \\ \hline \hline
\bf \multirow{3}{*}{A}  & \bf 1 & 2 Bi-LSTM & Mean & 0.9147 & 0.8162 & 0.6816  \\ 
                        & \bf 2 & 2 Bi-LSTM & Voting & 0.9116 & 0.8091 & 0.6696  \\ 
                        & \bf 3 & 2 Bi-LSTM & None & 0.9051 & 0.8029 & 0.6609  \\ \hline \hline
\bf \multirow{2}{*}{B}  & \bf 1 & 1 LSTM    & Mean & 0.7779 & 0.7567 & 0.6850  \\ 
                        & \bf 2 & 1 LSTM    & None & 0.7463 & 0.7306 & 0.6656  \\ 

\end{tabular}
\end{center}
\caption{\label{tab:official} Official submission results in different measures}
\end{table*}

In Table \ref{tab:official}, we show also results of submitted all models in 3 measures, micro F1, macro F1 and binary F1 (F1 score over positive samples). As an official results, binary F1 measure was taken. From these results, we can observe that model ensemble can significantly help obtain better results for both subtasks.

\subsection{Model ensemble results}
In this section, we discuss results of each model from model ensemble in detail for both subtasks. 

Table \ref{tab:ensemble_a} shows results of each model used for model ensemble for subtask A. We can observe that the best model obtained binary F1 score 0.6609 and both types of model ensembles get better results up to 2 percents than each model separately. For mean model ensemble first 3 models were used and for voting ensemble all 5 models were used.

\begin{table}[ht]
\begin{center}
\begin{tabular}{c|r|r|r}
\bf model & \bf micro F1 & \bf macro F1 & \bf binary F1  \\ \hline \hline
\bf  1 & 0.9051 & 0.8029 & 0.6609  \\ 
\bf  2 & 0.9050 & 0.8000 & 0.6550  \\ 
\bf  3 & 0.9075 & 0.8021 & 0.6577  \\ 
\bf  4 & 0.9099 & 0.8013 & 0.6543 \\
\bf  5 & 0.9159 & 0.8061 & 0.6601  \\ \hline
\bf  mean & 0.9147 & 0.8162 & 0.6816  \\
\bf  voting & 0.9116 & 0.8091 & 0.6696  \\ 
\end{tabular}
\end{center}
\caption{\label{tab:ensemble_a} Results of unsubmitted models in different measures for subtask A}
\end{table}

Table \ref{tab:ensemble_b} shows results of each model used for model ensemble for subtask B. We can see that the best model obtained binary F1 score 0.6770 and both types of model obtained better results than each model separately. For both types of model ensemble all 3 models were used. Results for voting ensemble were not part of the official submissions. 

\begin{table}[ht]
\begin{center}
\begin{tabular}{c|r|r|r}
\bf model & \bf micro F1 & \bf macro F1 & \bf binary F1  \\ \hline \hline
\bf  1 & 0.7742 & 0.7517 & 0.6770   \\ 
\bf  2 & 0.7730 & 0.7446 & 0.6593  \\ 
\bf  3 & 0.7463 & 0.7306 & 0.6656  \\ \hline
\bf  mean & 0.7779 & 0.7567 & 0.6850  \\ 
\bf  voting & 0.7574 & 0.7803 & 0.6830   \\ 
\end{tabular}
\end{center}
\caption{\label{tab:ensemble_b} Results of unsubmitted models in different measures for subtask B}
\end{table}

\subsection{Error analysis}

We provide also error analysis of proposed model for both subtasks. We made 3 official submissions for subtask A and 2 for subtask B. 

In Table \ref{tab:error}, we show simple results from confusion matrix for subtask A and also for subtask B. For subtask A, we can observe that the main problem of our proposed models was high number of false positive labels and our models predicted presence of suggestion too often. In subtask B, there is more problematic prediction of non-suggestion labels, where number of false negative samples is much bigger. This problem can be caused also due to different distributions in training and test datasets. We also used for training dataset from a different domain, which even highlighted this problem.
We used the same class weight modification for loss in subtask B as was used in subtask A.

\begin{table}[ht]
\begin{center}
\begin{tabular}{c|c|r|r|r|r}
\bf task & \bf submission & \bf TP & \bf FP & \bf FN & \bf TN  \\ \hline
\bf \multirow{3}{*}{A}  & \bf 1 & 76 & 60 & 11 & 686  \\ 
                        & \bf 2 & 75 & 62 & 12 & 684 \\ 
                        & \bf 3 & 77 & 69 & 10 & 677 \\ \hline \hline
\bf \multirow{2}{*}{B}  & \bf 1 & 199 & 34 & 149 & 442  \\
                        & \bf 2 & 208 & 69 & 140 & 407 \\ 
\end{tabular}
\end{center}
\caption{\label{tab:error} Error analysis for submissions }
\end{table}

\subsection{Unsubmitted models}

In this section, we present results of models, which were not used to make an official submission. We experimented with these models for subtask A and also subtask B. Results can be found in Table \ref{tab:unsub_a} and \ref{tab:unsub_b}. Each model in this section is used without model ensemble and we can compare results with the best models themselves. In each table, best model indicates best submitted model without any model ensemble.

The only modification used for model 1 is replacing ELMo word representation with BERT. Obtained word representations from pre-trained BERT performed much worse than ELMo representation. This fact was also observed while evaluating on development (trial) dataset.

Model 2 had a more significant modification, where LSTM encoder was replaced with transformer network \cite{Vaswani2017}. Due to high memory requirements of this model we were not able to run full encoder of the original network and used only smaller part with 6 layers and 4 head attention layers.

As we observed in error analysis of submitted results (see Table \ref{tab:error}), one of the significant problems was predicting too many positive labels. For all submissions, we used re-balancing class weights for loss function based on distribution in train dataset (0.6625, 2.0384). In model 3, we changed class weights to be more balanced (1.0 and 2.0). In model 4, we used completely balanced weights (1.0 and 1.0) and in model 5, we tried to change class weights to prefer negative labels (2.0 and 1.0).

\begin{table}[ht]
\begin{center}
\begin{tabular}{c|r|r|r}
\bf model & \bf micro F1 & \bf macro F1 & \bf binary F1  \\ \hline
\bf  best & 0.9051 & 0.8029 & 0.6609  \\ 
\bf  1 & 0.8415 & 0.7214 & 0.5384  \\ 
\bf  2 & 0.9123 & 0.7952 & 0.6404  \\ 
\bf  3 & 0.9314 & 0.8440 & 0.7272  \\ 
\bf  4 & 0.9459 & 0.8577 & 0.7457  \\ 
\bf  5 & 0.9495 & 0.8479 & 0.7236  \\ 
\end{tabular}
\end{center}
\caption{\label{tab:unsub_a} Results of unsubmitted models in different measures for subtask A}
\end{table}

For subtask B, we also experimented with use of pre-trained weights from language model. We trained language model on dataset of hotel reviews -- arguana \cite{Wachsmuth2014review}. Unfortunately, without fine-tuning on in-domain dataset used for classification, this model did not obtained better results. We show results of this experiment as model 1 in Table \ref{tab:unsub_b}. We suppose that further experiment would be needed with combination of class re-balancing for loss and also fixing pre-trained weights. Since our language model was trained with GloVe embeddings, we had to use GloVe also in training for this task. Models 2 and 3 show results with change of class weight for loss function to prefer positive labels (1.0, 4.0 and 1.0, 5.0).

\begin{table}[ht]
\begin{center}
\begin{tabular}{c|r|r|r}
\bf model & \bf micro F1 & \bf macro F1 & \bf binary F1  \\ \hline
\bf  1 & 0.7208 & 0.6698 &  0.5401  \\ 
\bf  2 & 0.7961 & 0.7844 & 0.7341  \\ 
\bf  3 & 0.8264 & 0.8186 & 0.7810  \\ 

\end{tabular}
\end{center}
\caption{\label{tab:unsub_b} Results of unsubmitted models in different measures for subtask B}
\end{table}

As we showed in this section, our further experiments along with error analysis showed also significant improvement in comparison to performed official submissions. We believe further work can provide even better results, especially in combination with model ensemble. 

\section{Conclusions}

We proposed a neural model architecture for suggestion mining. For subtask A, we employed bidirectional LSTM encoder, which consisted from 2 stacked layers followed by self attention. For subtask B, better performance proved only one layer in one direction to reduce learning process and over-fitting to train domain. Our experiments showed that pre-trained ELMo word representations performed much better than pre-trained BERT. We also performed other experiments with different setups of our architecture, which were not submitted as official results. As we showed, model ensemble can significantly improve results compared when using only single models. 

To obtain better results, we would need to employ transfer learning to a much bigger extent, especially for subtask B. We could also consider further experiments with re-balancing of class weights for loss function, as we predicted too many suggestions, especially for subtask A. Another possible experiments would employ transformer network, which we were not able to fully employ due to high resource requirements. Interesting would be also employing some pattern approaches, which proved as very successful for subtask B in baseline provided by organizers. Code for our submission can be found in GitHub repository\footnote{https://github.com/SamuelPecar/NL-FIIT-SemEval19-Task9}.

\section*{Acknowledgments}

This work was partially supported by the Slovak Research and Development Agency under the contracts No. APVV-17-0267 and
No. APVV SK-IL-RD-18-0004,  
and by the Scientific Grant Agency of the Slovak Republic grants No. VG 1/0725/19 
and No. VG 1/0667/18. The authors would like to thank for financial contribution from the STU Grant scheme for Support of Young Researchers.


\end{document}